\documentclass[journal]{IEEEtran}
\usepackage{tikz}
\usepackage{subfigure}
\usepackage{subfloat}
\usepackage{multirow}
\usepackage{amssymb}
\usepackage{amsmath}
\usetikzlibrary{trees}
\graphicspath{{./}}

\ifCLASSINFOpdf
\else
\fi
\usepackage{amsmath}

\begin{document}
%
\title{A PCB Dataset for Defects Detection and Classification}
%
%
%

\author{{Weibo Huang}, 
        {Peng Wei}
\thanks{Weibo Huang, Peng Wei are with the Key Laboratory of Machine Perception, Shenzhen Graduate School, Peking University, Shenzhen, CN, 518052  {weapon}@pku.edu.cn}
\thanks{J. Doe and J. Doe are with Anonymous University.}}

%
%

\markboth{Journal of \LaTeX\ Class Files,~Vol.~14, No.~8, August~2018}%
{Shell \MakeLowercase{\textit{et al.}}: Bare Demo of IEEEtran.cls for IEEE Journals}
%



\maketitle

\begin{abstract}
To coupe with the difficulties in the process of inspection and classification of defects in Printed Circuit Board (PCB), other researchers have proposed many methods. However, few of them published their dataset before, which hindered the introduction and comparison of new methods. In this paper, we published a synthesized PCB dataset containing 1386 images with 6 kinds of defects for the use of detection, classification and registration tasks. Besides, we proposed a reference based method to inspect and trained an end-to-end convolutional neural network to classify the defects. Unlike conventional approaches that require pixel-by-pixel processing, our method firstly locate the defects and then classify them by neural networks, which shows superior performance on our dataset.
\end{abstract}


%
\IEEEpeerreviewmaketitle

\section{Introduction}
%
%
%
%

Printed circuit board (PCB) is the fundamental carrier in electronic devices on which a great number of elements are placed. The quality of the PCB will directly impact the performance of electronic devices. To avoid the shortcoming of manual detection, easily being fatigued, low efficiency, for instance, automated optical inspection (AOI) based on machine vision has been widely used in industry. As PCB becomes more and more complicated, the tasks of detection and classification defects are also more difficult than before. Currently, there are few public datasets on the Internet on PCB, many methods proposed in published papers used their own images, which is not convenient for other researchers to compare their new methods. For the purpose of solving above problems, we produced a public colorized synthesized PCB dataset with defects that is available to other people who want to design and evaluate their approaches.  

\begin{figure}[!t]
	\centering
	\includegraphics[width=0.8\linewidth]{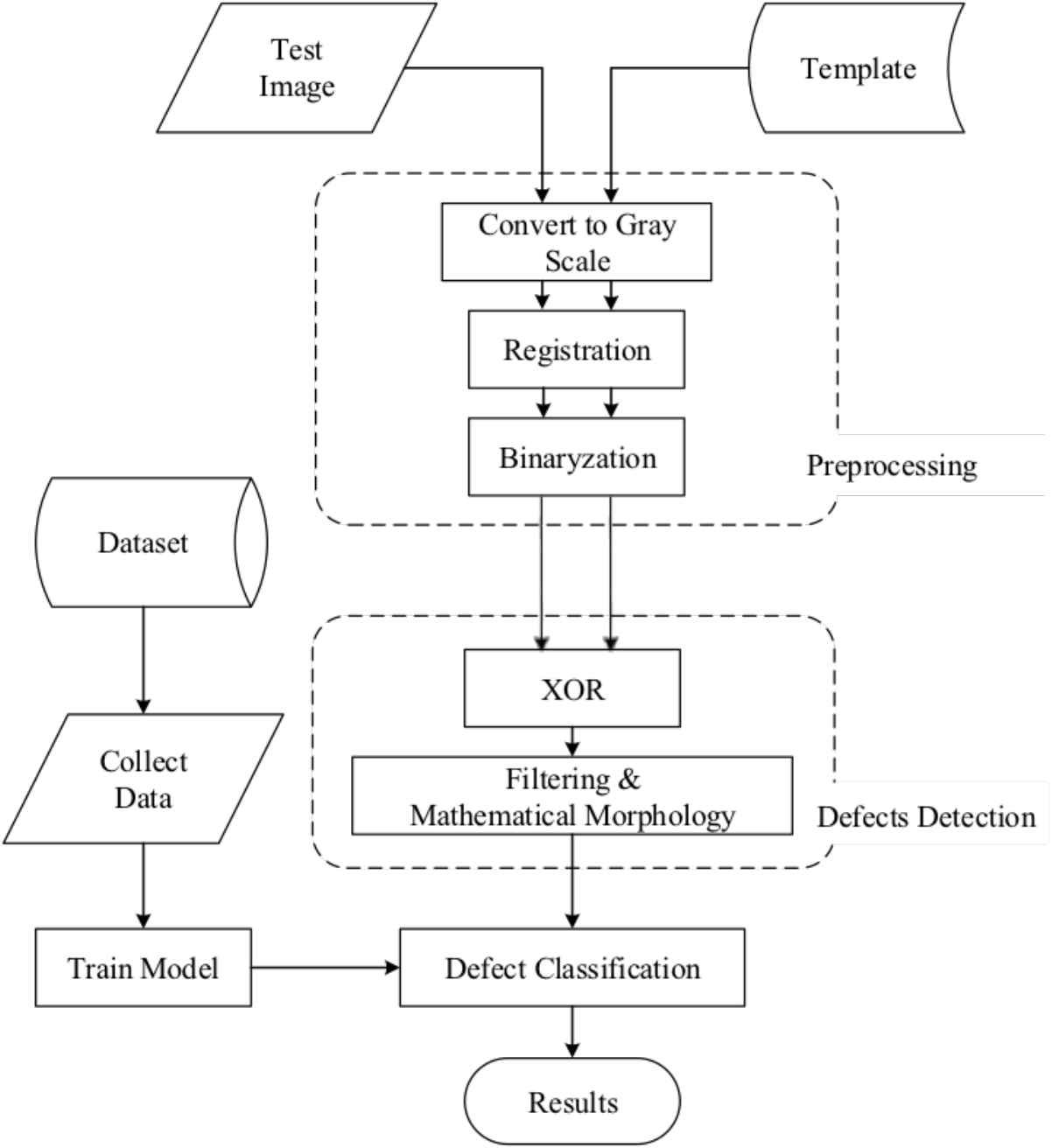}
	\caption{The flow chart for PCB inspection. Test image and template will be separately preprocessed and compared to locate defects, then these located defects will be send into trained neural network model to get results.}
	\label{fig_flow_cahrt}
\end{figure}%

Conventional AOI methods for inspecting printed circuit board can be divided into 3 main streams \cite{Wu1996} : reference comparison approach, non-reference verification approach, and hybrid approach. In reference comparison approach, a standard image which is called template will be prepared firstly, and then a PCB need to be inspected will be compared with the template to find the unknown defects. Though it is straightforward and easy to use, there are also many factors that we have to take in consideration, unbalanced illumination, inaccurate registration, vast storage requirements, etc. In the non-reference verification approach, the aim of the method is to find out if wiring track, pad and hole are in the compliance with design without a template board. This approach does not have the limits of reference method, nevertheless, it may have difficulties in detecting large defects. In the hybrid approach, reference method and non-reference method are combined, this approach will have the merits of the two basic methods, meanwhile, it requires high computation capacity.

Some various methods have been proposed on this task. Wen-Yen Wu et al. \cite{Wu1996} introduced the development of an automated visual inspection system for PCB. It utilized an elimination-subtraction method which directly subtracts the template image from the inspected image, and then conducts an elimination procedure to locate defects in the PCB. Each detected defect is classified by three indices: the type of object detected, the difference in object numbers, and the difference in background numbers between the inspected image and the template. LI Zheng-ming et al. \cite{Zheng2012Detection} also used digital image processing technology based reference method to classify the defects by getting the number of connected regions, euler numbers, area of defects of the template and inspected image respectively. The result of experiment showd that the method can achieve automatic real-time detection. Vikas Chaudhary et al. \cite{Chaudhary2018} listed 14 kinds of defects that belong two types: positive, negative, and segmented the image into 3 parts: wiring tracks, soldering pads and holes. Each defect can be classified by comparing pixels, number of connected components in the corresponding part. Shashi Kumar et al. \cite{Kumar2017} proposed a non-referential based approach in consideration of the difficulties in registration. In his work, inspected image was segmented into copper and non-copper parts to analyze separately, and a 3D color histogram was utilized to capture the global color distribution. The effectiveness of this model is evaluated on real data from PCB manufacturing industry and accuracy is compared with previously proposed non-referential approaches. Rudi Heriansyah et al. \cite{Heriansyah2003} introduced a new technique that is to classify the defects using neural network paradigm. Various defective patterns representing corresponding defect types were designed and thousands of defective patterns had been used for training and testing. The result showed the effectiveness of defect classification technology based on neural network.

Because of the intuitiveness, simpleness, and the development of computer hardware and algorithms, reference comparison method is used to inspect defects in our approach. In addition, convolutional neural network highlights outstanding performance in computer vision tasks, like classification, object detection, segmentation, etc. Therefore, in the defect classification task, we do not search the features of the image, instead, we introduce an end-to-end neural network to classify the inspected defect regions. The experimental results prove its effectiveness. The flow chart of the whole experiment process is shown in Figure \ref{fig_flow_cahrt}. 
     
Before our work, there are some public datasets on Printed circuit board assembly (PCBA) \cite{Pramerdorfer2015} which is a kind of board after all the components and parts have been soldered and installed on the PCB and can accomplish the electronic function it was designed for. Inspection of PCBA is for the purpose of recycling when the PCBA is eliminated, however, it is not appropriate for us to use these PCBA datasets because our target is naked PCB that has no components. In this paper, we present a synthesized dataset that consists of 1386 naked PCB images. Half of them are in right orientation as templates with different defects and other half are manually rotated to simulate the situation when PCBs are not correctly placed. All the images originate from 10 standard template boards which are checked by human. Each PCB has 3 to 5 defects and we provide corresponding bounding box for every defect. Rotation information are also provided for PCBs rotated. Many PCB-related methods on detection, classification and registration problems can be conducted on this dataset, and various methods can be compared as well. The dataset is free available online\footnote[1]{www.baidu.com} .

The paper is organized as follows, section \uppercase\expandafter{\romannumeral1} introduces the backgrounds of PCB dataset and main stream methods on defects detection and classification. section \uppercase\expandafter{\romannumeral2} details the procedure of image acquisition, labeling and defect statistics. Our reference comparison based method and the detection part is given in section \uppercase\expandafter{\romannumeral3}. section \uppercase\expandafter{\romannumeral4} presents convolutional neural network based model that we use to classify defects and the experiment results. Conclusions is shown in section \uppercase\expandafter{\romannumeral5}.

\begin{figure}[!t]
	\centering
	\includegraphics[width=2.5in]{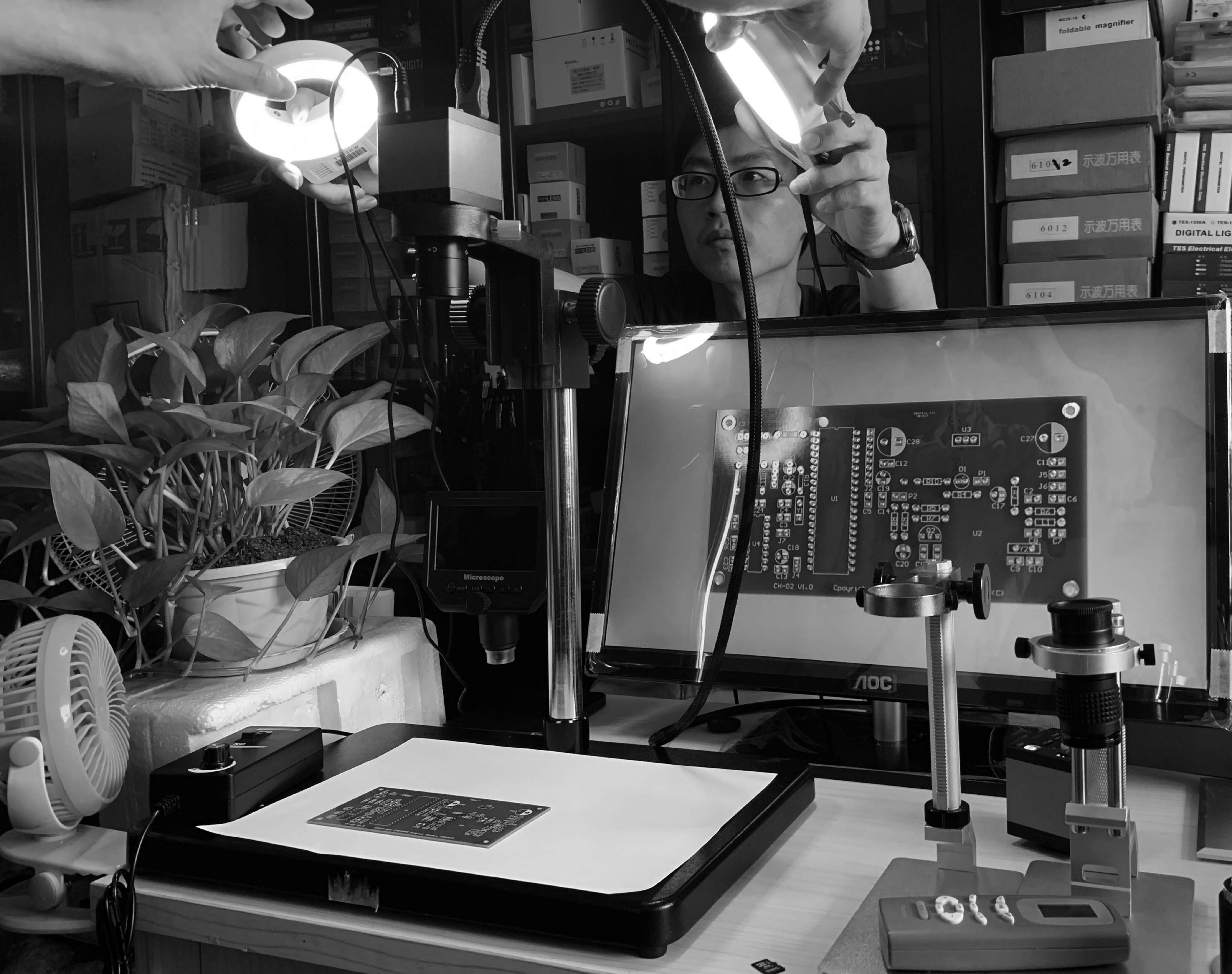}
	\caption{The PCB image acquisition system consisting of light source, workbench, support, camera and image process unit.}
	\label{fig_workbench}
\end{figure}%

\section{Image acquisition and statistics}
In addition to procedure and equipment related to image acquisition and dataset production, some statistics on the dataset is arranged in this section.
\subsection{Image acquisition}
To ensure the representativeness of the dataset, we build a PCB image acquisition system that resembles the practical AOI system used in inspection process, as is shown in Figure \ref{fig_workbench}.

The image of template board is captured by a 16-megapixel HD industrial camera equipped with CMOS sensor, and it can be controlled by computer software or a remote control. In order to adapt to different PCB sizes and avoid edge distortion, an undistorted zoomable industrial lens is also mounted, the focal length can be adjusted between 6-12mm and the maximum aperture is f1.6. Light source is also a key part of AOI, to avoid specular reflection of the board, possible shadows and minimize the effects of uneven illumination on subsequent steps, two frosted ring LED source equipped with special diffuse matting board are introduced to effectively overcome the adverse effects of illumination. The resolution of original photo is 4608$ \times $3456 pixels, which will be adjusted according to the size of each board when make defects. 
 
\begin{figure}[htbp]
\centering
\subfigure[PCB with mouse bite.]{
\begin{minipage}[b]{.5\textwidth}
  \centering
  \includegraphics[width=0.8\linewidth]{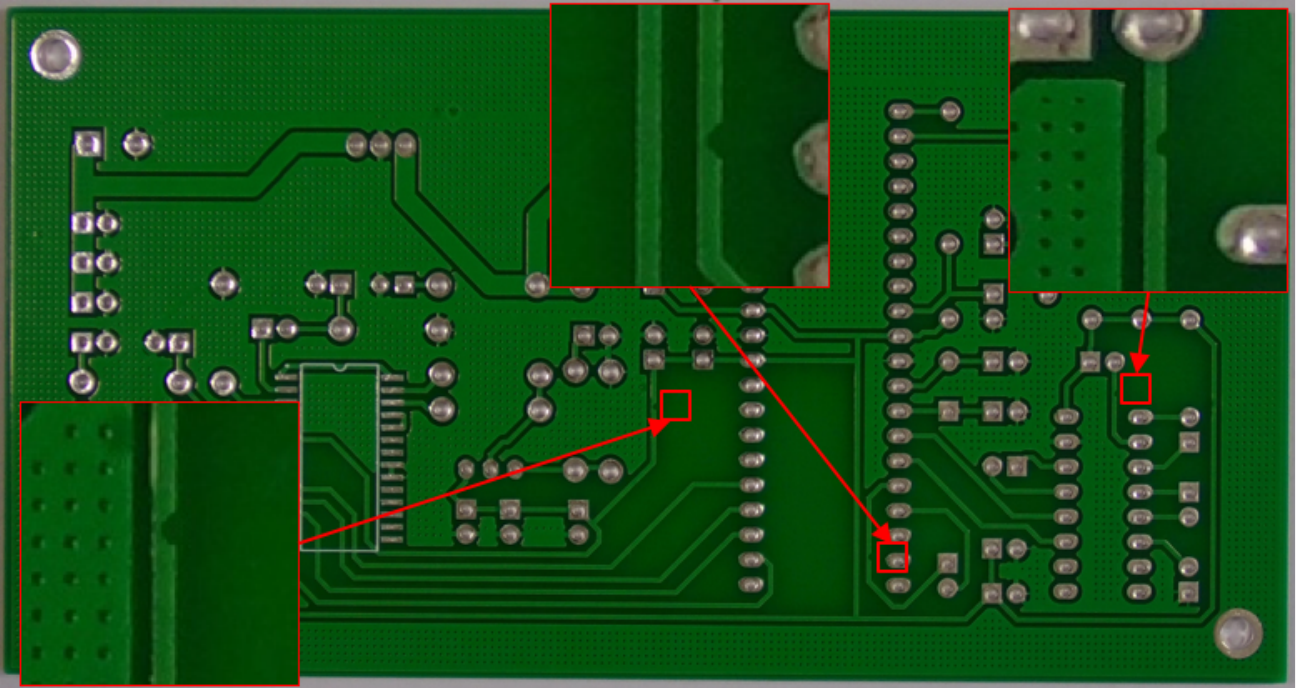}
\end{minipage}%
}

\subfigure[PCB with mouse bite and rotation.]{
\begin{minipage}[b]{.5\textwidth}
  \centering
  \includegraphics[width=0.8\linewidth]{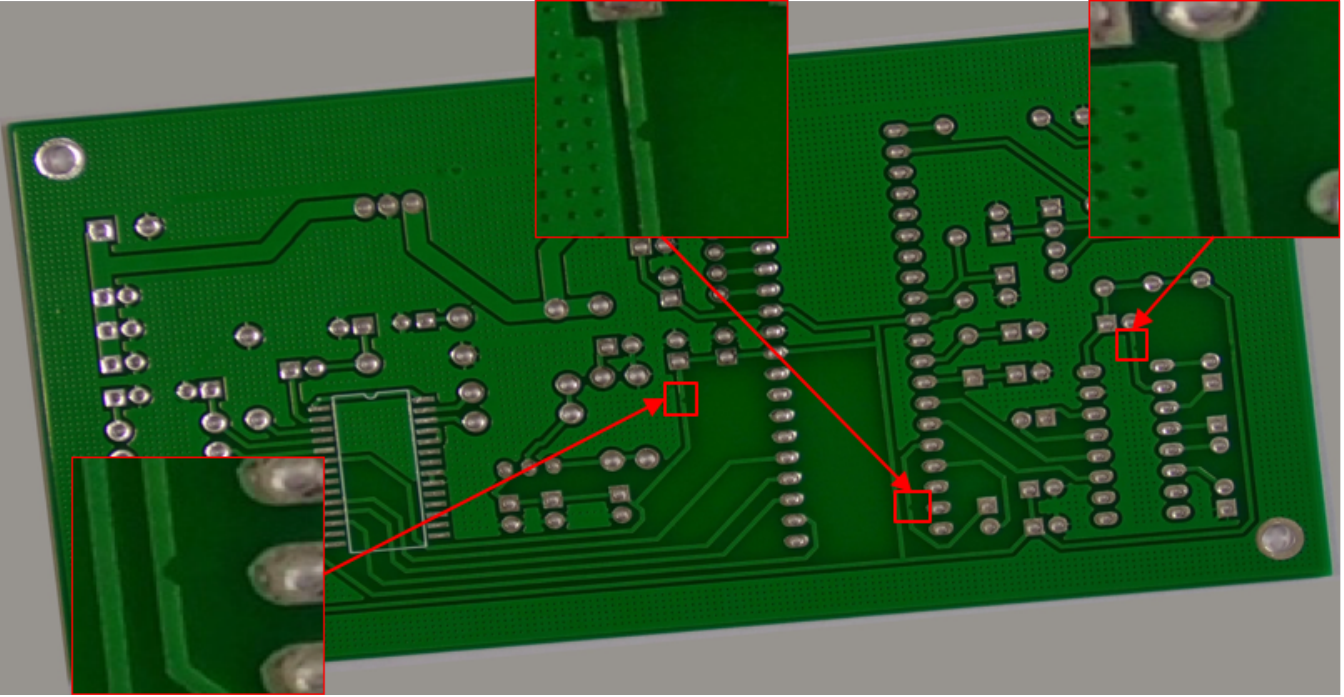}
\end{minipage}
}
\caption{Samples of the PCB with defects in the dataset, (a) is the defects image with the same position as template, (b) is the image with random orientation.}
\label{fig_dataset}
\end{figure}

After getting cropped image, we make 6 types of defects by photoshop, which is a graphics editor published by Adobe Systems. The defects we defined are: missing hole, mouse bite, open circuit, short, spur, spurious copper. Each image in the dataset has 3 to 5 defects of the same category in different places. Besides, we provide bounding box and coordinate information for every defect in every image, which is convenient for other researchers to know where the defect is. On some inspection platforms, PCB can be fixed by mechanical devices to maintain good position. However, on the assembly line, without fixing equipments, the position and the angle of the test PCB in the taken photo may distinguish from each other. Given this circumstance, in addition to the defects images with the same position as the templates, we also provide images with random orientations to represent the situation where the image is not appropriately placed in practical detection process. The angular difference between each image and the corresponding template image is also given so that the designing and evaluating of registration algorithm could be implemented on these images, the samples of the dataset can be seen in Figure \ref{fig_dataset}. 

\subsection{Statistics}
\begin{figure}
\centering
\tikzstyle{every node}=[draw=black,thick,anchor=west]
\tikzstyle{selected}=[draw=red,fill=red!30]
\tikzstyle{optional}=[dashed,fill=gray!50]
\begin{tikzpicture}[%
  grow via three points={one child at (0.5,-0.7) and
  two children at (0.5,-0.7) and (0.5,-1.4)},
  edge from parent path={(\tikzparentnode.south) |- (\tikzchildnode.west)}]
  \node {DATASET}
    child { node {Annotations}
      child { node {Missing\_{}hole}}
      child { node {.....}}
      }
    child [missing] {}				
    child [missing] {}			
    child { node {Images}
      child { node {Missing\_{}hole}}
      child { node {.....}}
    }
    child [missing] {}				
    child [missing] {}	
    child { node [selected] {PCB\_{}USED}
      child { node {01.JPG}}
      child { node {02.JPG}}
      child { node {.....}}
    }
    child [missing] {}				
    child [missing] {}
    child [missing] {}				
    child { node {rotation}
      child { node {Missing\_{}hole}}
      child { node {Missing\_{}hole\_{}angles.txt}}
      child { node {.....}}
    }
    child [missing] {}				
    child [missing] {}				
    child [missing] {};
\end{tikzpicture}
	\caption{Structure of the dataset, the whole dataset is divided into 4 parts.}
	\label{fig_dataset_structure}
\end{figure}
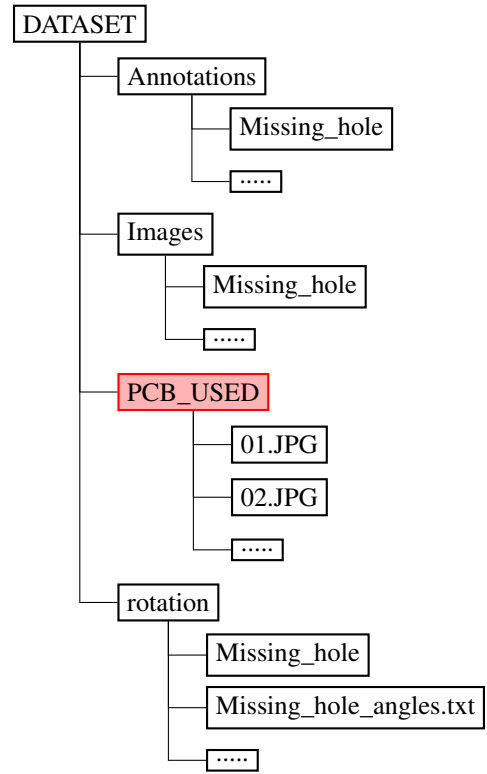

The dataset has four main parts, which are placed in four different folders. The Images folder stores the PCB photos with the same position as the templates, and all the photos of a defect type are put in a folder of the same name. Information of bounding boxs of each image is kept in a .xml file that saved in Annotations folder. PCB\_{}USED folder contains the 10 template images we used in the dataset. Moreover, rotation folder has PCB images with orientations, and rotation angles are also placed with image names in .txt files in this folder. The structure of the dataset is shown in the tree diagram in Figure \ref{fig_dataset_structure}.
  
\vspace{1ex}

\begin{figure}[!t]
	\centering
	\includegraphics[width=.8\linewidth]{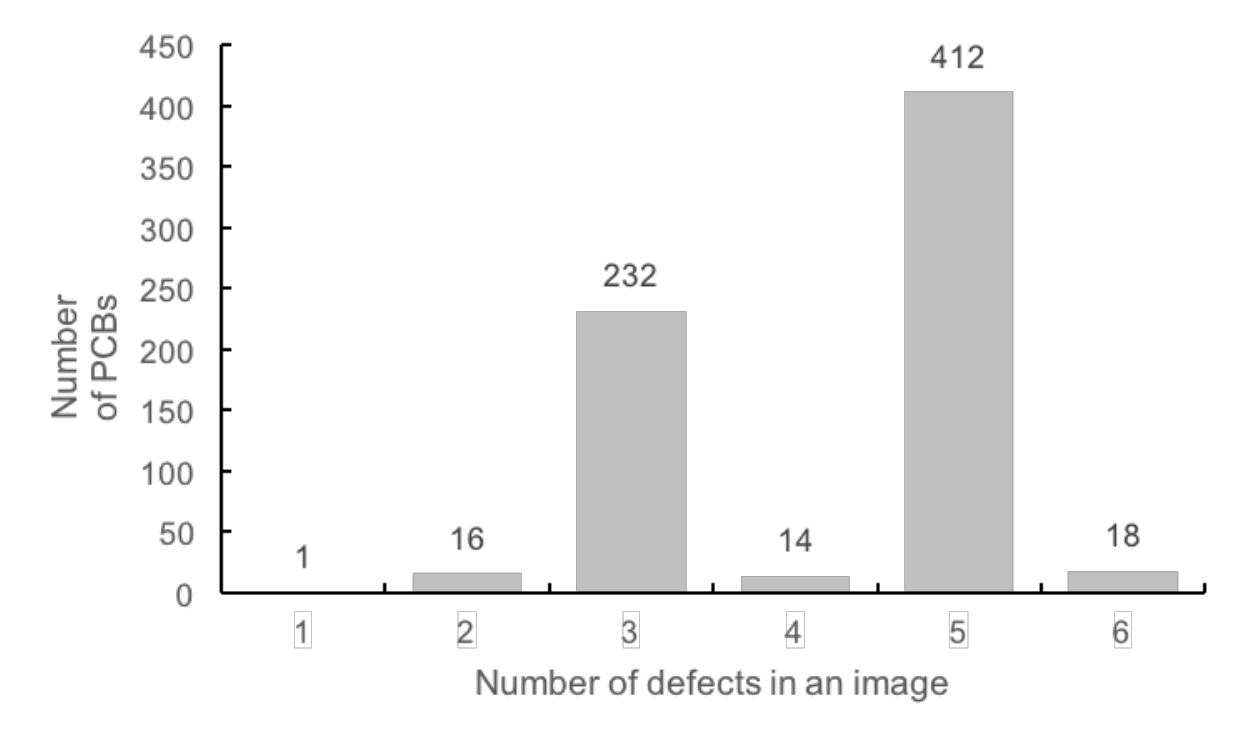}
	\caption{The distribution of defects per PCB.}
	\label{fig_dis_def}
\end{figure}

\begin{figure}[!t]
	\centering
	\includegraphics[width=.8\linewidth]{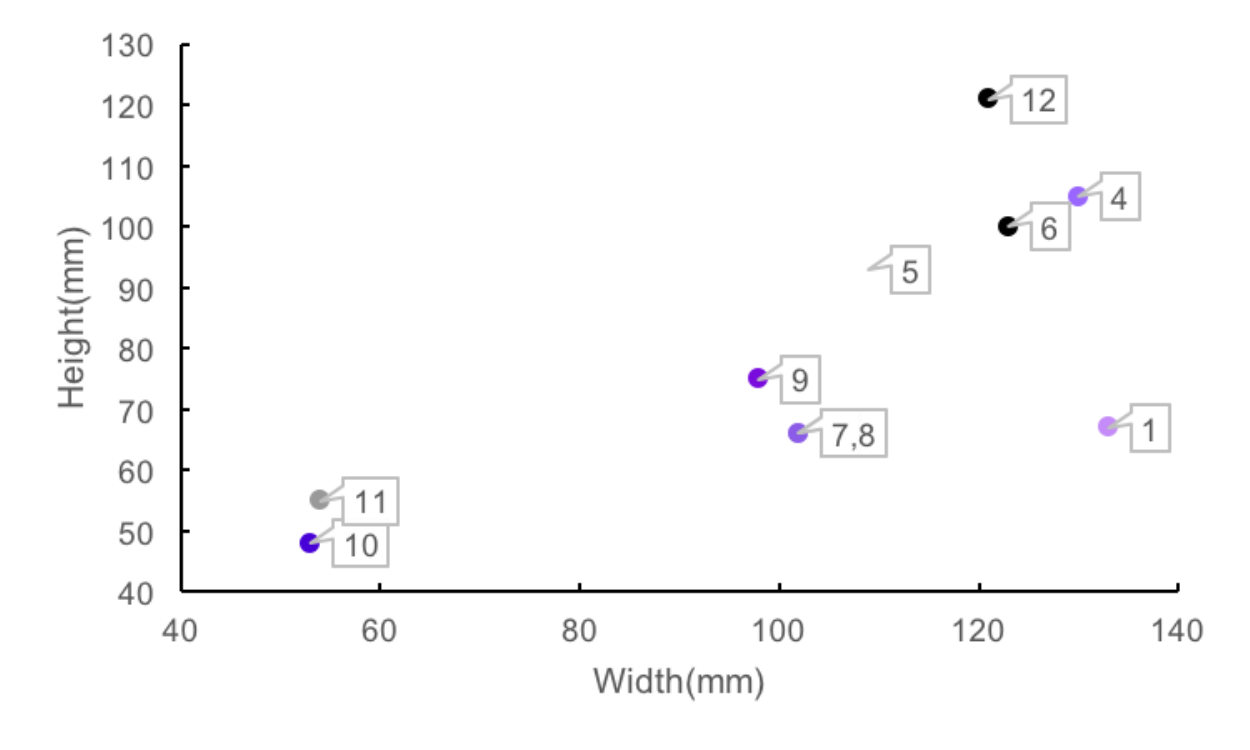}
	\caption{The statistics of height and width of the templates.}
	\label{fig_height_with}
\end{figure}

Details of the figures for PCB images and defect samples are listed in table \ref{table_dis_def}. Figure \ref{fig_dis_def} shows the distribution of defects per PCB. It is visible that the majority of PCBs have less than 6 defects, and most PCBs have more than 2 defects. In Figure \ref{fig_height_with}, the height and width of every template is given. We can see that the largest PCB size in the dataset is $20mm \times 120mm$, while the minimum is $53mm \times 48mm$. In order to facilitate the use of our dataset, we provide API for easy access in python, the .py file in the dataset will intuitively show the bounding box of each defect in the dataset.

%

\begin{table}[!t]
\renewcommand{\arraystretch}{1.3}
\caption{The figures for PCB and defect samples (listed in the brackets).}
\label{table_dis_def}
\centering
\begin{tabular}{c||c}
\hline
Category & Number(defects) \\
\hline
Missing hole & 115(497) \\
\hline
Mouse bite &  115(492)\\
\hline
Open circuit & 116(482)\\
\hline
Short &  116(491) \\
\hline
Spur &  115(488) \\
\hline
Spurious copper & 116(503)\\
\hline
\end{tabular}
\end{table}

%

%

\section{Preprocessing and detection}
In this section, prepocessing steps like registration and binaryzation are applied, followed by XOR and mathematical morphology operation that helped to loacte defects. 
\subsection{Registration}
\begin{figure}[htbp]
	\centering
	\subfigure[Feature points in tempalte.]{

			\includegraphics[width=0.44\linewidth]{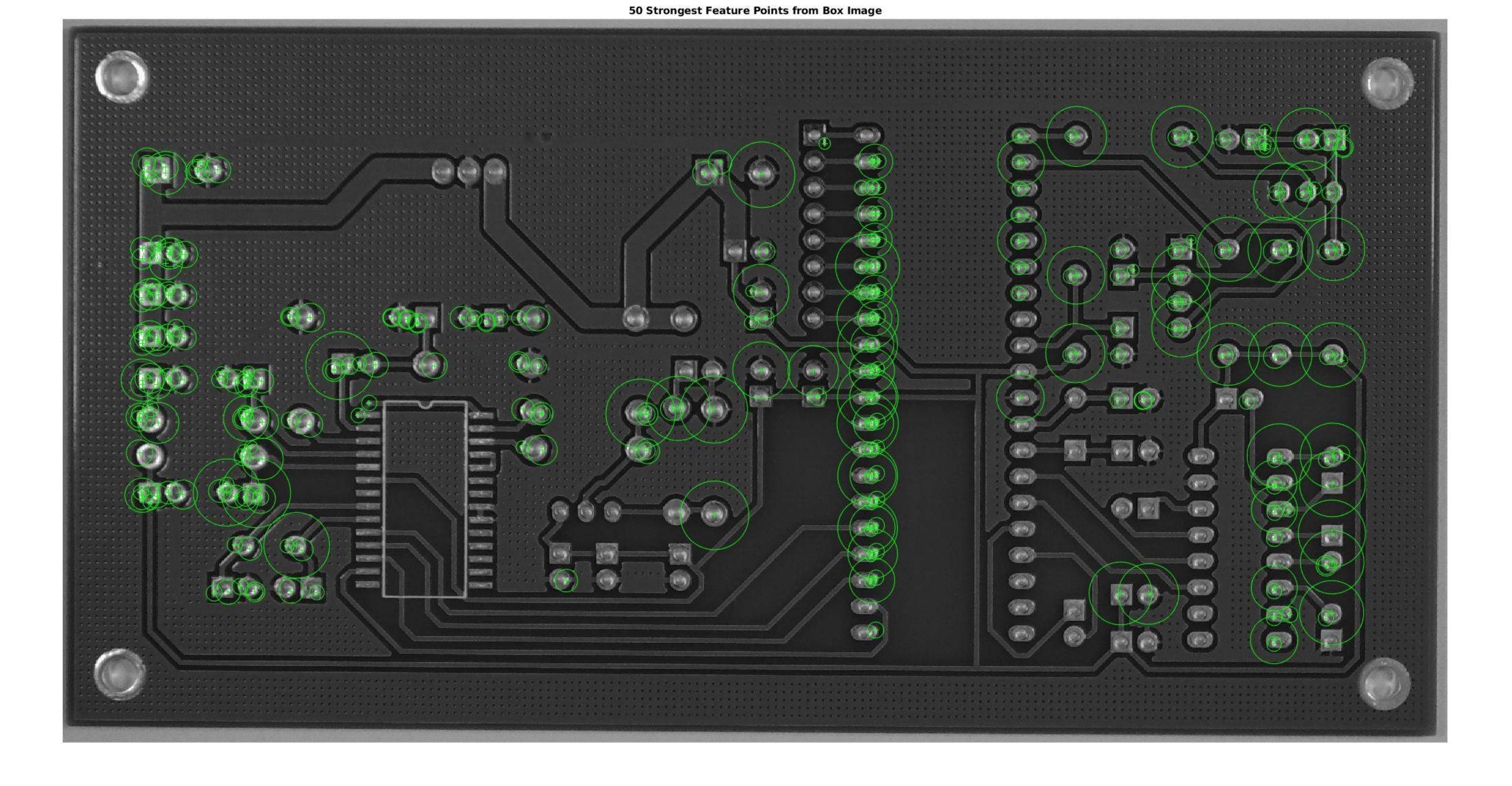}
	
	}
	\quad 
	\subfigure[Feature points in test image.]{

			\includegraphics[width=0.44\linewidth]{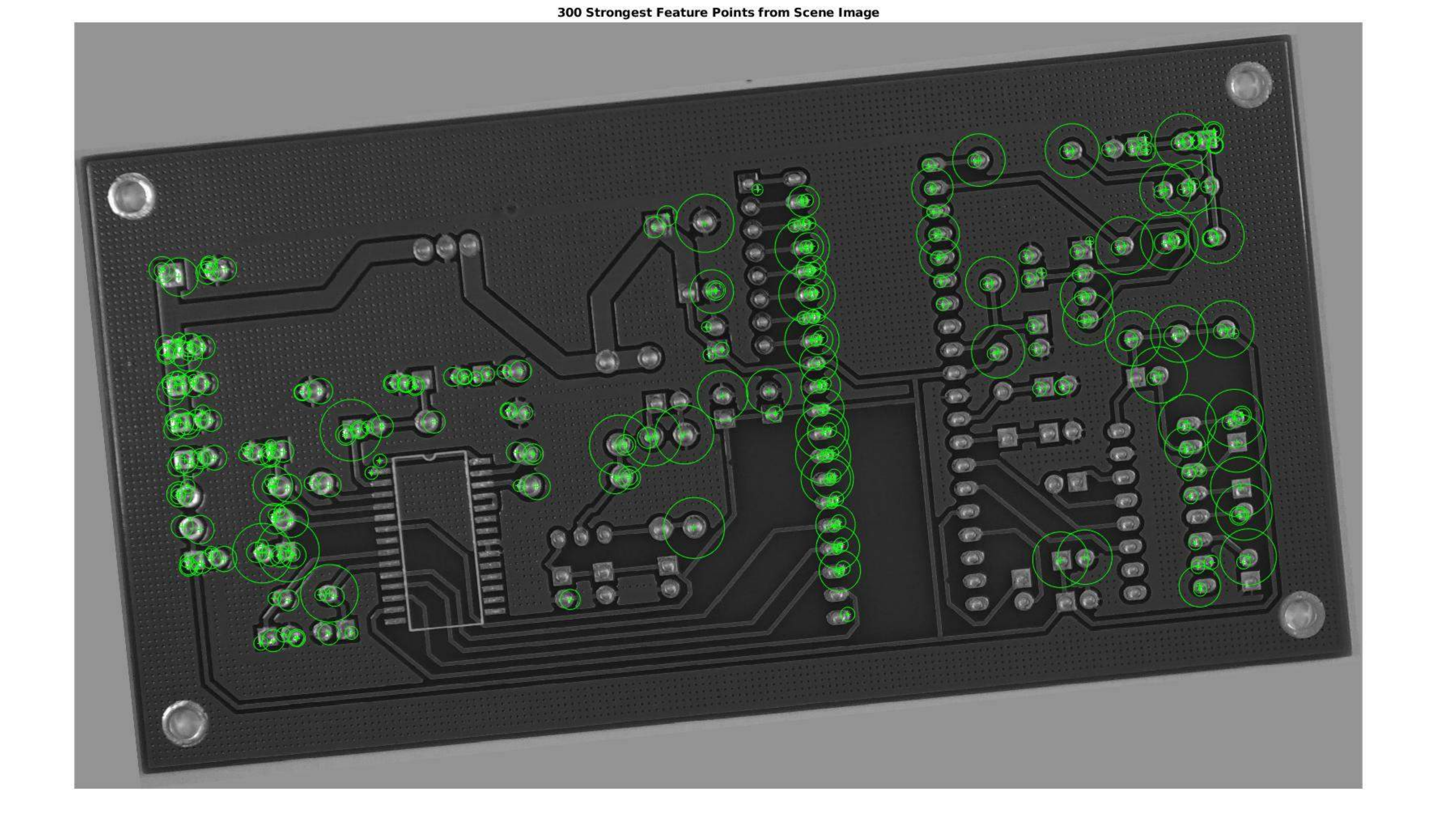}
	
	}
	
	\subfigure[Matched fearure points.]{

		\includegraphics[width=0.9\linewidth]{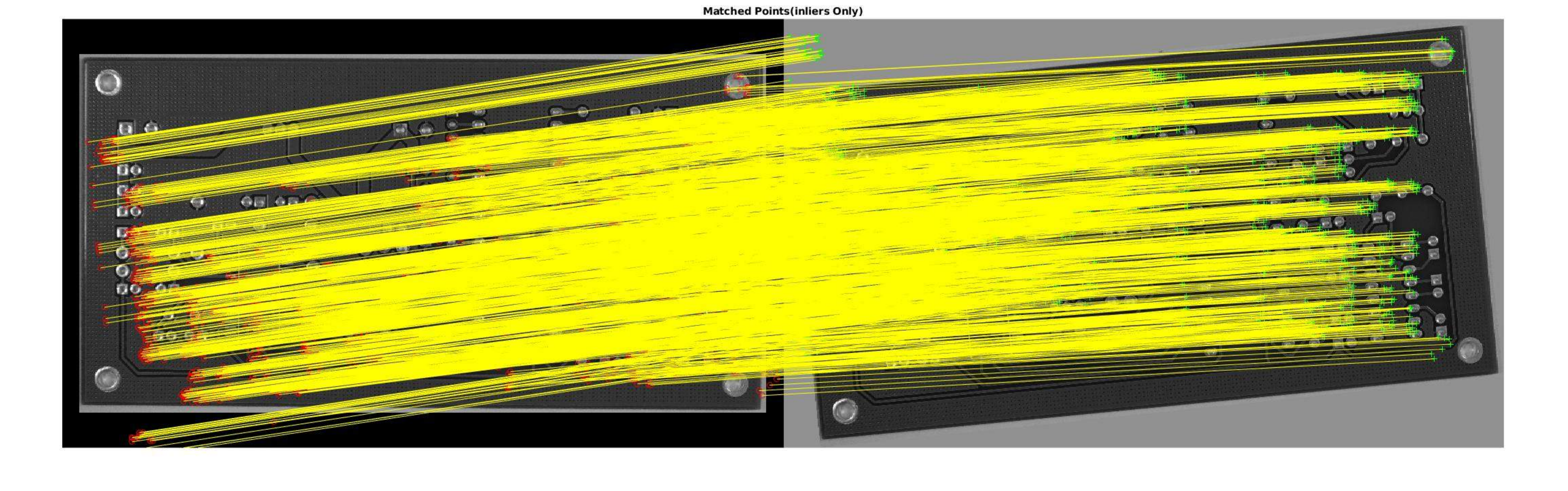}
		
	}
	
	\subfigure[Transformed test image.]{

		\includegraphics[width=0.6\linewidth]{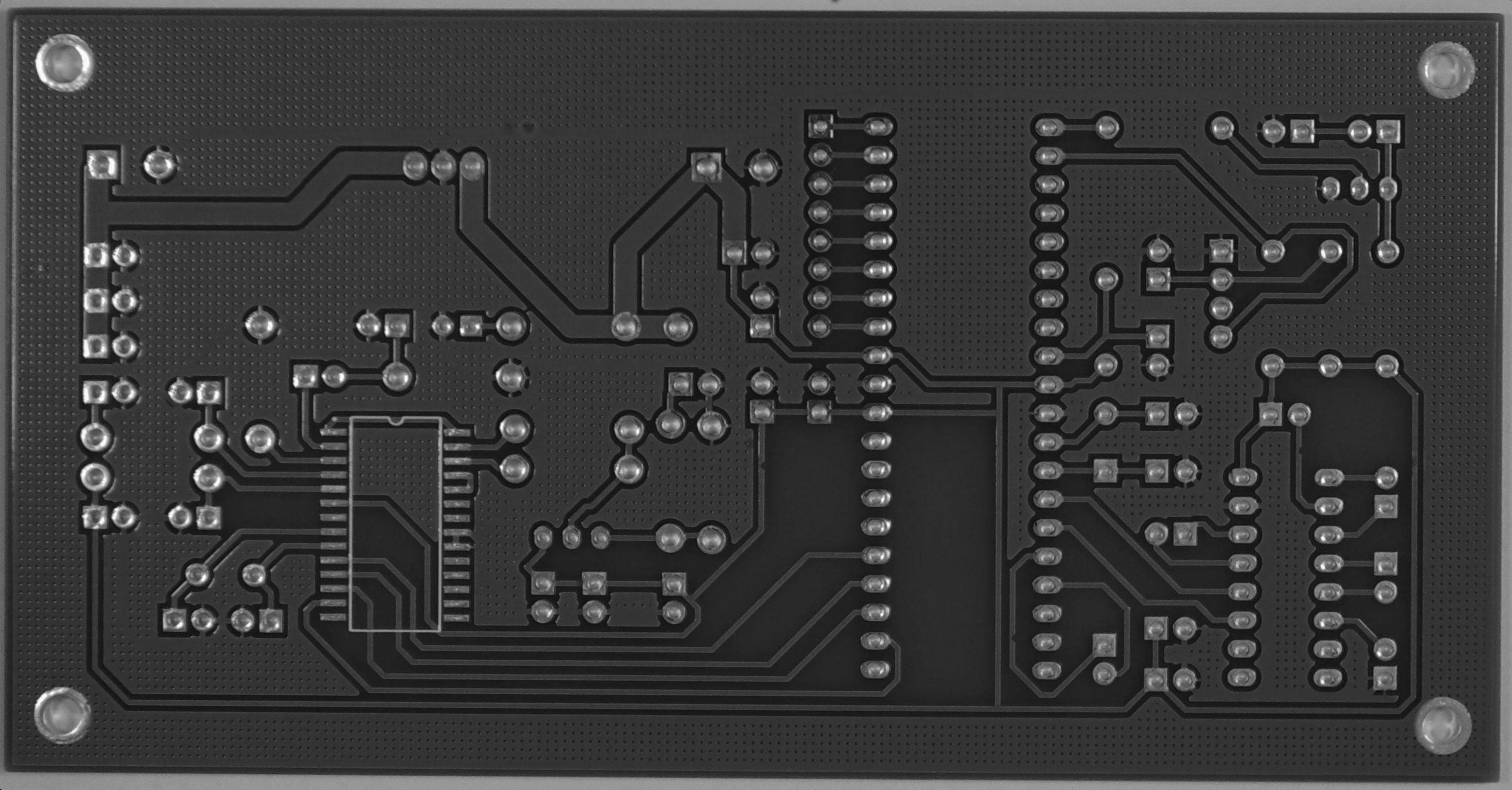}
			
	}
	\caption{The selected and matched feature points and transformed test image in our experiment.}
	\label{fig_match}
\end{figure}

The printed circuit board are placed on workbench or assembly line while photographing, which would result in the differences in direction and geometric center between PCB to be inspected and template board. So registration is indispensable in reference comparison based method. A test image and template image will be converted into gray image first, then feature points of the two images are extracted and matched, finally the transformation matrix is calculated to transform the test image into the same orientation and position as the template image. In this paper, Speeded Up Robust Features (SURF) \cite{Bay2006SURF} algorithm is used to extract feature points in PCB. It is an improvement of Scale Invariant Feature Transform (SIFT) \cite{Lowe2004Distinctive}, with less computational complexity, and can run faster compared with SIFT. The feature points selected by SURF and SIFT are both stable and they are rotation, scale, luminance invariant. Although SIFT has better matching effect than SURF in the case of scale and rotation transformation, SURF has better matching effect under brightness change, considering the practical application scenarios, SURF is chosen for PCB registration. Once get the SURF feature points of the tempalte and test image, a 2-D geometric transform will be estimated from matching points and the test image will be recovered by the geometric transform. Figure \ref{fig_match} shows the selected and matched feature points and transformed test image in our experiment. 
         

\subsection{Binaryzation}
\begin{figure}[htbp]
	\centering
	\subfigure[Binary image of template.]{
		\begin{minipage}[b]{.5\textwidth}
			\centering
			\includegraphics[width=0.8\linewidth]{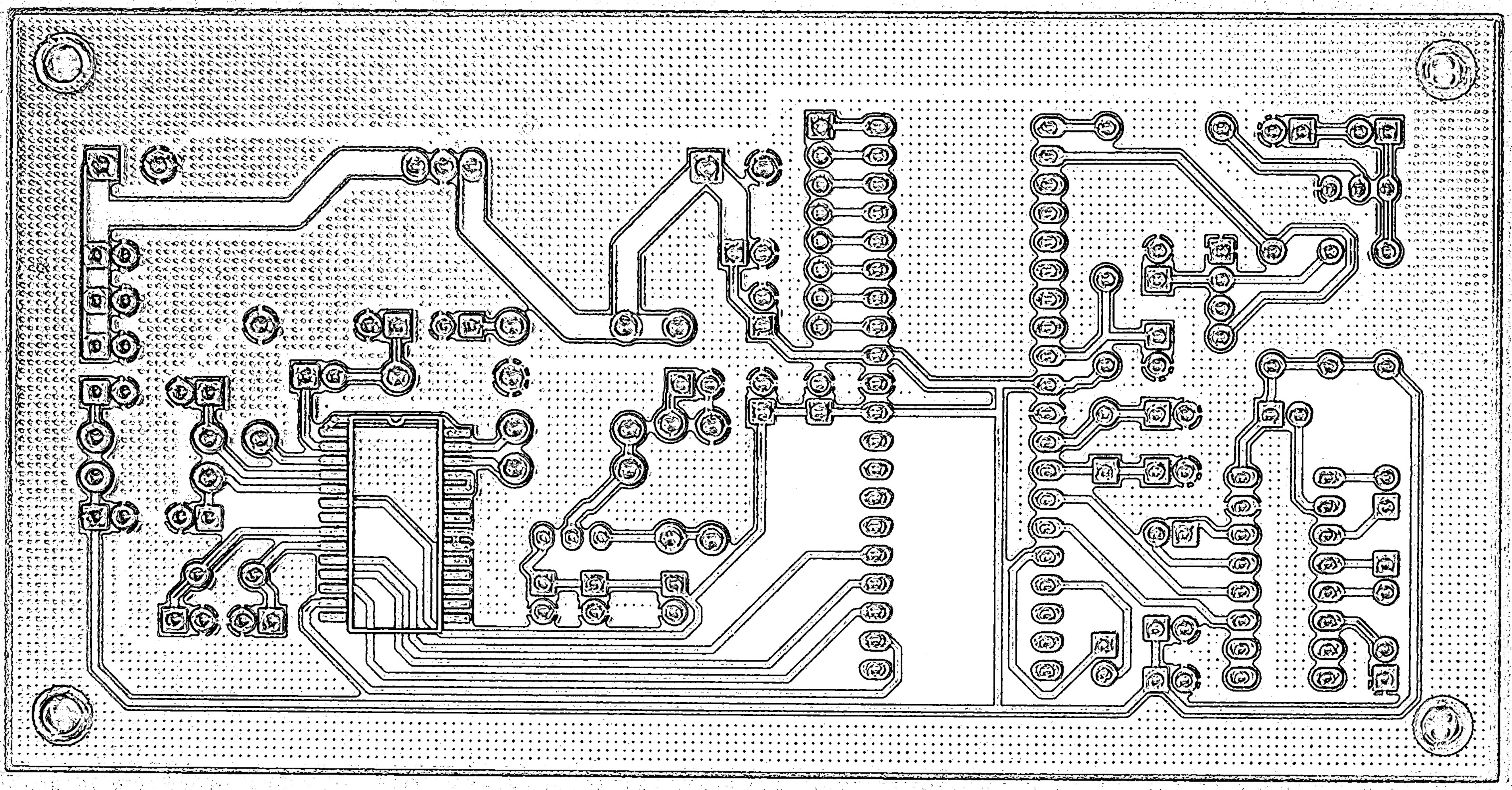}
		\end{minipage}%
	}
	
	\subfigure[Binary image of test image.]{
		\begin{minipage}[b]{.5\textwidth}
			\centering
			\includegraphics[width=0.8\linewidth]{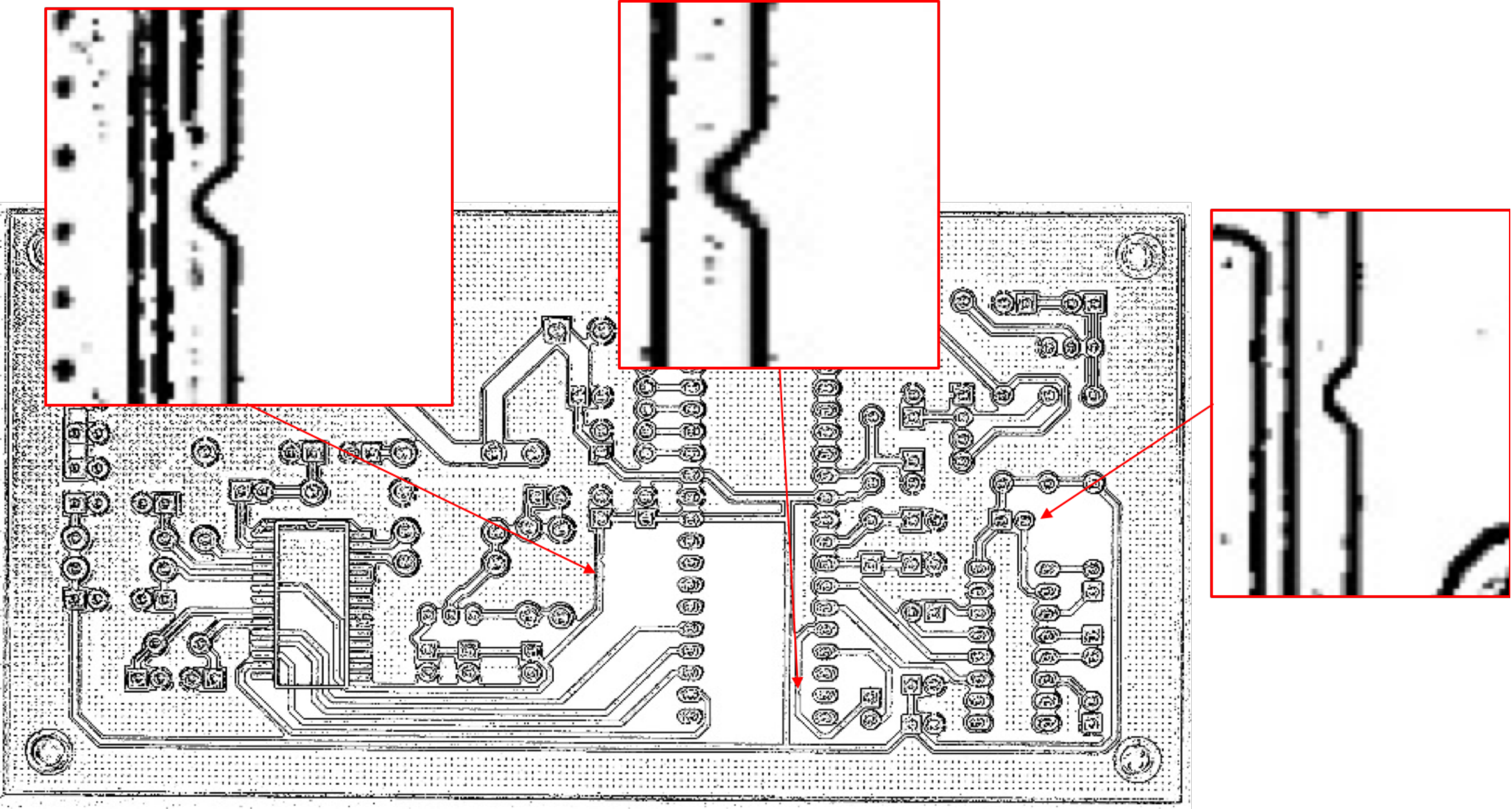}
		\end{minipage}
	}
	\caption{The binary images coming from our dataset. It is clear that defects are outlined in the binary map.}
	\label{fig_binaryzation}
\end{figure}

It is not easy to directly compare two colorized or gray-scale images due to the fact that they are easily influenced by illumination. Nevertheless, by using a binary map, the outline and shape of the PCB are only expressed in black and white, which is more convenient for comparison. The next step is to convert gray-scale image into binary image to get the location of defects. There are many methods for image binaryzation, in our method, adaptive threshold segmentation algorithm \cite{Gonzalez2009Digital} is chosen. Instead of using a global value as threshold value, adaptive threshold algorithm calculates thresholds for small regions of the image, because PCB image may have different lighting conditions in various areas. For every pixel $(x,y)$, the threshold value $T(x,y)$ is the weighted sum of $blocksize \times blocksize $ neighbourhood values where weights are a gaussian window. The gaussian kernel is defined as follows:
\begin{equation}
\displaystyle
G_{i}=\alpha \times e^{\frac{{-\frac{i-(ksize-1)}{2}}^{2}}{2 \times sigma^{2}}}
\end{equation}
where $i=1, 2, ... , ksize-1 $ and $\alpha$ is the scale factor chosen so that $\sum G_{i}=1$, $ksize$ indicates aperture size and it should be odd, $sigma$ is gaussian standard deviation computed from $ksize$. Once $T(x,y)$ is calculated individually for each pixel in every region, the output value $dst(x,y)$ is defined in Equation 2: 

\begin{equation}
\displaystyle
dst(x,y)=\left\{\begin{matrix}
maxValue & if \quad src(x,y)>T(x,y)\\ 
0		 & otherwise 
\end{matrix}\right.
\end{equation}

where $maxValue$ is a non-zero value assigned to the pixels for which the condition is satisfied, usually set as 255. The bianry images coming from our dataset is shown in Figure \ref{fig_binaryzation}. 

\subsection{Localization of defects}
The result binary image is obtained by XOR binary image of template and tested image, the formula of XOR operation is defined in Equation 3:
\begin{equation}
\displaystyle
dst(I)=src1(I) \oplus src2(I) 
\end{equation}
where $dst(I)$ is the result binary image, $src1(I)$, $src2(I)$ are template binary map and tested binary map respectively. In the XOR, if the pixel values in the corresponding positions of the template and test image are the same, the pixel value of the position in result image will be 1 after XOR, if not, the result value will be 0.

However, the result binary image may contain a great number of noises and unwanted pseudo defects. To get real defects, median filtering \cite{Huang1979A} and mathematical morphological \cite{Dr1999Morphological} processing are used. Median filtering is a non-linear filtering technique used to eliminate tiny noise points in the image. The basic idea is to sort the pixel values of the neighborhood of a pixel point $(x,y)$, and take the intermediate value to replace the value of the original pixel.
Morphological processing is a theory and technique for the analysis and processing of geometrical structures, the basic morphological operators are erosion, dilation, opening and closing which are defined in Equation 4, 5, 6, 7 continuously.

\begin{equation}
\displaystyle
A \ominus B=\left\{z\mid B_{z}\subseteq A \right\}
\end{equation}
where $B_{z}=\left\{b+z \mid b\in B\right\}$. In erosion, if the structuring element B has a center, then the erosion of A by B can be understood as the locus of points reached by the center of B when B moves inside A. Generally speaking, erosion can make the range of the target area smaller, which can be used to eliminate small and meaningless objects in an image.

\begin{equation}
\displaystyle
A \oplus B=\left\{z\mid (\overline{B})_{z} \cap A \neq \varnothing \right\}
\end{equation}
where $\overline{B}=\left\{x \mid -x \in B\right\}$. In dilation, if B has a center on the origin, then the dilation of A by B can be understood as the locus of the points covered by B when the center of B moves inside A. The dilation can be used to make the target boundary to expand outward to fill in some holes and eliminate small particle noises existing in the target area.

\begin{equation}
\displaystyle
A \circ B= (A \ominus B)\oplus B
\end{equation}
The opening of A by B is obtained by the erosion of A by B, followed by dilation of the resulting image by B, which will remove isolated points, burrs and bridges, while the overall position and shape of the target area remain unchanged. 

\begin{equation}
\displaystyle
A \bullet B =(A \oplus B)\ominus B
\end{equation}
The closing of A by B is obtained by the dilation of A by B, followed by erosion of the resulting structure by B. It can fill the small holes and close the small cracks, keeping the overall position and shape unchanged as the opening. 

\begin{figure}[htbp]
	\centering
	\subfigure[Result image after XOR operation.]{
		\begin{minipage}[b]{.5\textwidth}
			\centering
			\includegraphics[width=0.8\linewidth]{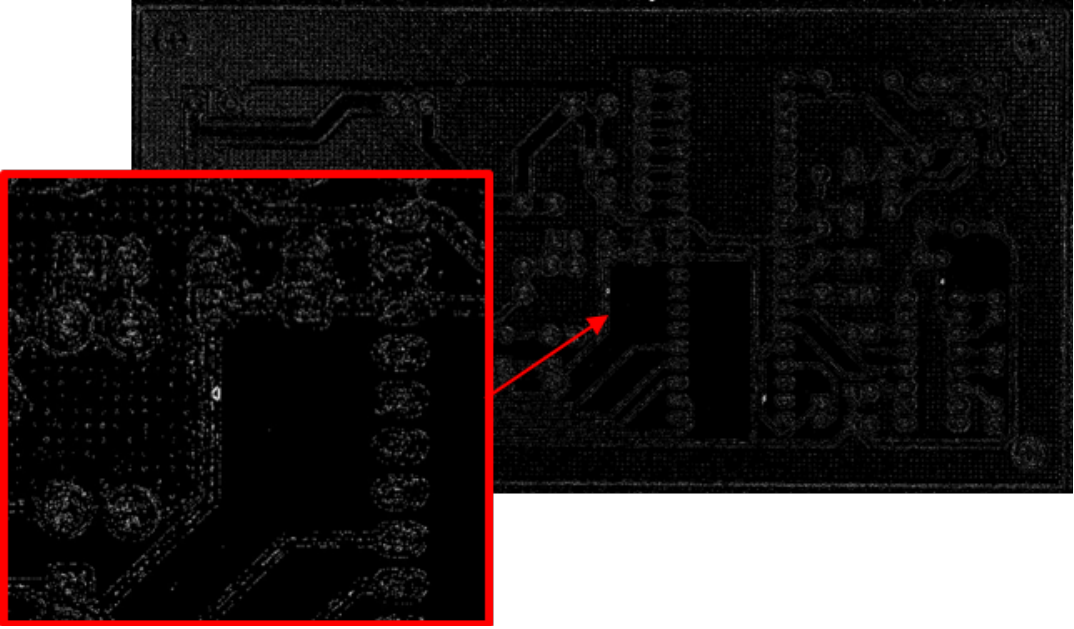}
		\end{minipage}%
	}
	\centering
	\subfigure[Defect image after filtering and morphology operation.]{
		\begin{minipage}[b]{.5\textwidth}
			\centering
			\includegraphics[width=0.8\linewidth]{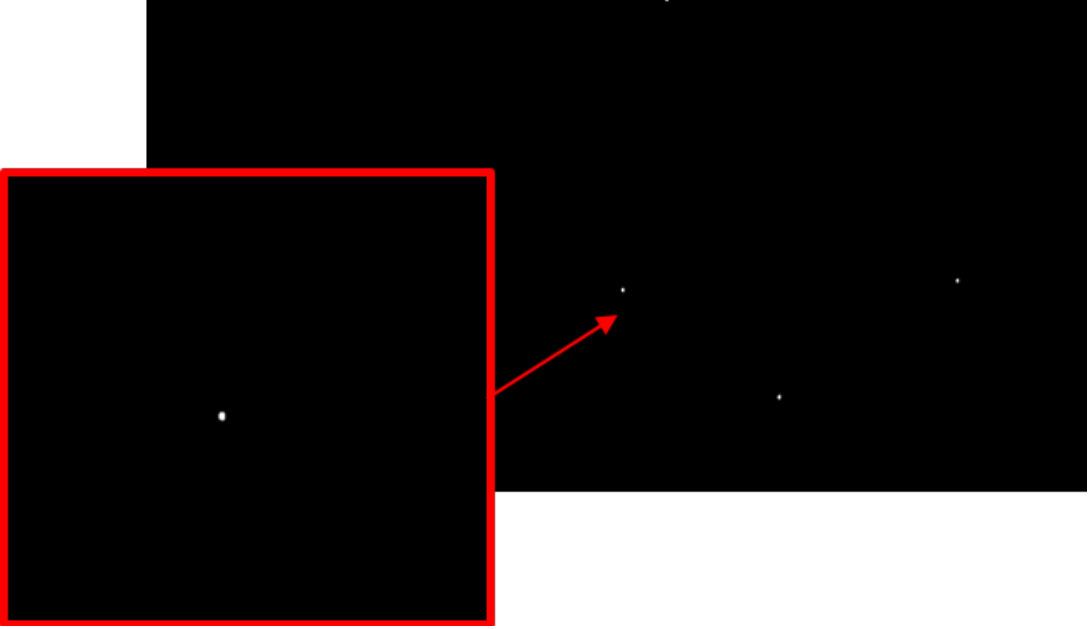}
		\end{minipage}
	}
	\caption{Result image after XOR operation (a) and defect image after filtering and mathematical morphology operation (b).}
	\label{fig_fil_math}
\end{figure}

In this paper, the result image of XOR operation is filtered by a $5 \times 5$ kernel first to get rid of some small isolate points, then closing operation with $15 \times 15 $ rectangle element is taken so that local parts of defects would be connected and enhanced, followed by a opening operation with $ 3 \times 3 $ rectangular element. The main object in a binary image will be highlighted by using closing and opening operation continuously. In addition, we continue to set the area threshold to remove too small points, followed by setting non-maximum value suppression (NMS) that will remove adjacent redundant candidate regions. All the steps on filtering and mathematical morphological processing in this paper is shown in Table \ref{table_step_fil_mm}. Final result image is pure without other points except for the area that real defects locate. In this case, the location of defects can be obtained from the connected areas, the result image after XOR operation and defect image after filtering and mathematical morphology operation are drawn in Figure \ref{fig_fil_math}. 

\section{Experiment and classification}
In this section, experiment based on convolutional neural network for defect classification is introduced, including data preparation, model selection and analysis of experiment results.  
\subsection{Preparing for data}
\begin{figure}[!t]
	\centering
	\includegraphics[width=.6\linewidth]{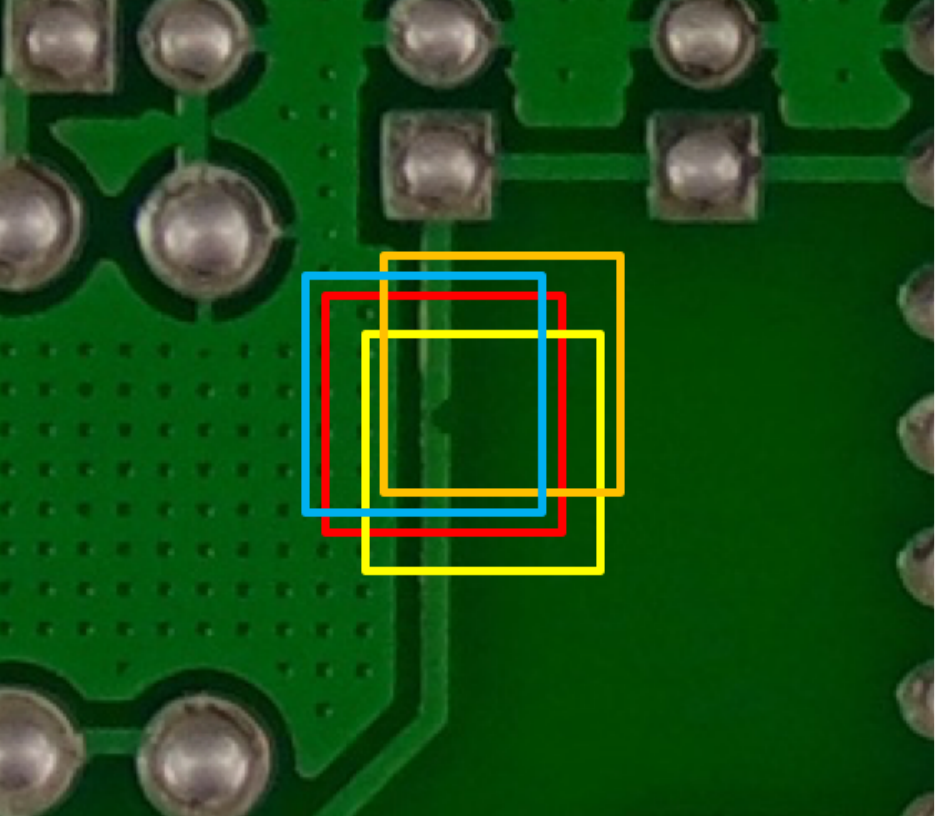}
	\caption{The depicts of data augmentation, the red frame is original bounding box labeled in dataset, and the other 3 frames (blue, yellow, orange) are created for producing more defect images.}
	\label{fig_strategy}
\end{figure}

After getting the location of defects, the next step is to identify the defect category. Conventional methods are based on pixel-by-pixel comparison between template and test image to select enough features to represent defects \cite{Wu1996,  Zheng2012Detection,  Chaudhary2018,  Heriansyah2003}, which would have non-ideal result if the binaryzation is in poor condition. Nevertheless, by using an end-to-end deep learning model, the defect image can be sent to the model as input directly to obtain a classification result, thereby avoiding extracting pixel-based features from the binary image. In this paper, a convolutional neural network based method is utilized to classify defects. The priori task for training and testing neural network is to prepare enough data. Considering that bounding box in our dataset has already given the coordinates of each defect, we cut the image in the bounding box as the data for neural network. In order to have data augmentation to produce more training images, we change the position of defects in the image by randomly make 5 pixel to 10 pixel offset on the existing coordinates , as the Figure \ref{fig_strategy} shows. 

\begin{figure}[!t]
	\centering
	\includegraphics[width=.8\linewidth]{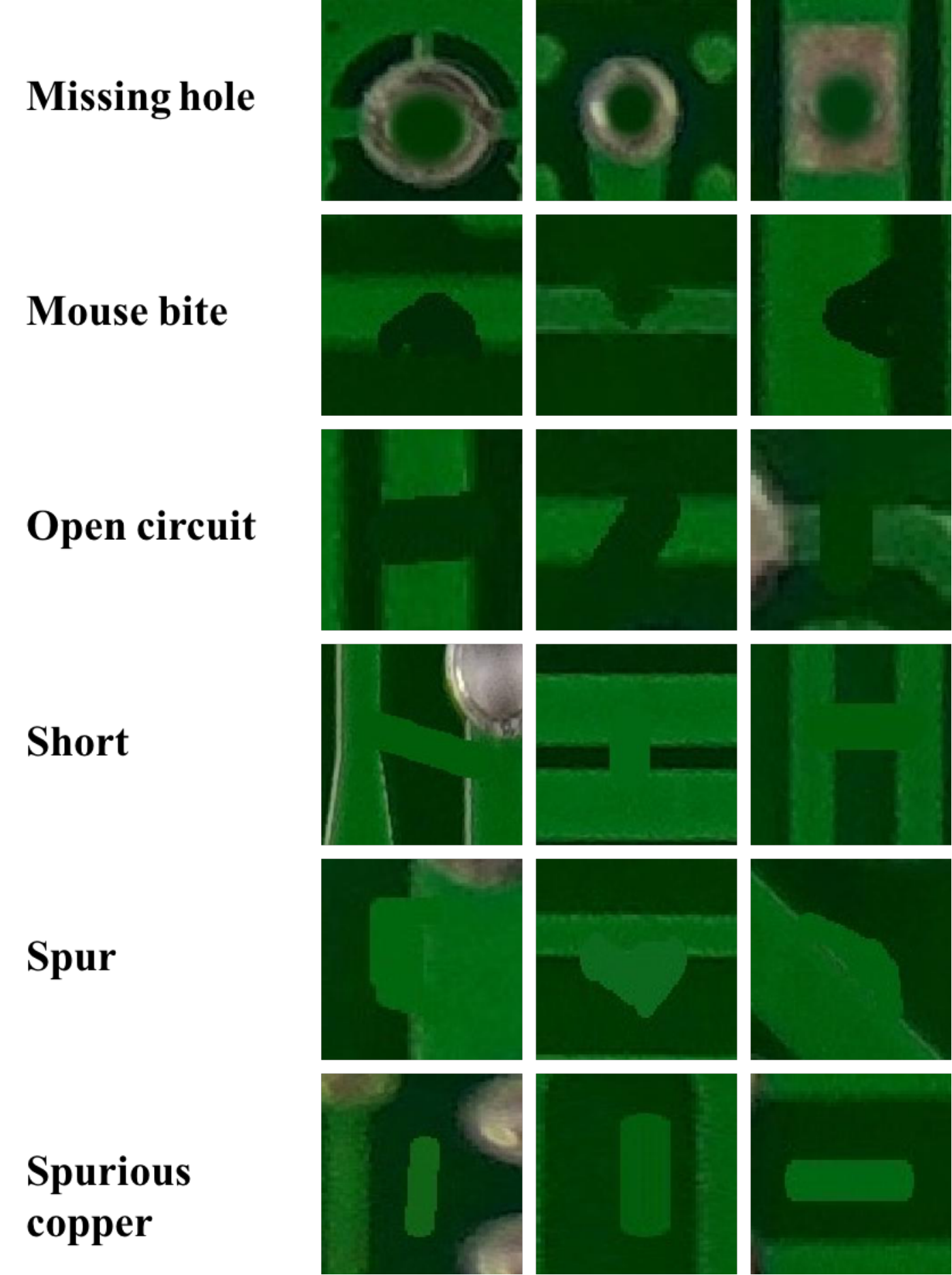}
	\caption{An example of the training data for neural network. All the resolutions of the images are resized to $64 \times 64$.}
	\label{fig_example_data}
\end{figure}

In this way, the size of data will be expanded and the generalization ability of the model will be enhanced. Resolution of each original defect image cropped from PCB dataset varies from one image to another. To facilitate the use of defect data, all images are resized to a resolution of $64 \times 64$, which are divided into 3 folders: train, val and test. Further, there are 6 sub-folders under each folder including all the images of the 6 defect types. An example diagram of the training data is shown in Figure \ref{fig_example_data}, and the size and distribution of the data is displayed in Table \ref{table_dis_def_for_train}.    
              
\begin{table}[!t]
	\renewcommand{\arraystretch}{1.3}
	\caption{Distribution of defects in training, val and test folders.}
	\label{table_dis_def_for_train}
	\centering
	\begin{tabular}{c||c||c||c}
		\hline
	  ~               & Train & Val  & Test \\
	  \hline
	  Missing hole    & 599   & 203  & 192  \\
	  \hline
	  Mouse bite      & 600   & 190  & 194  \\
	  \hline
	  Open circuit    & 598   & 189  & 177  \\
	  \hline
	  Short           & 600   & 189  & 177  \\
	  \hline
	  Spur            & 600   & 187  & 189  \\
	  \hline
	  Spurious copper & 600   & 203  & 203  \\
	  \hline
	  Total           & 3597  & 1161 & 1148 \\
	  \hline
	\end{tabular}
\end{table}

\subsection{Models}
Convolutional neural network (CNN) has powerful ability to extract features in pictures, and it has been widely used in many computer vision tasks such as classification \cite{inproceedings,Huang2017Densely} , segmentation \cite{article}, object detection \cite{Liu2016SSD}, etc. Recently, in the field of defect inspection, a lot of methods based on convolutional neural network have been adopted \cite{Faghih2016Deep, article1_ADCCTS, article_CSDSS}. The results showed their superiorities compared with conventional approaches. As tasks become more and more complicated, the CNNs also become increasingly deep to make sure that more features would be extracted to contribute to the final result. However, another problem called gradient diffusion occurs when gradient flow back to the beginning if the network is so deep. In this case, one common solution for the problem above is creating shortcut from early layers to later layers. In our method, inspired by Densenet \cite{Huang2017Densely}, to utilize the densely connection structure, a small and efficient network is designed to handle PCB defect classification problem.

%

\begin{figure*}[htbp]
	\centering
	\subfigure[Block with 6 layers.]{

			\centering
			\includegraphics[width=0.8\linewidth]{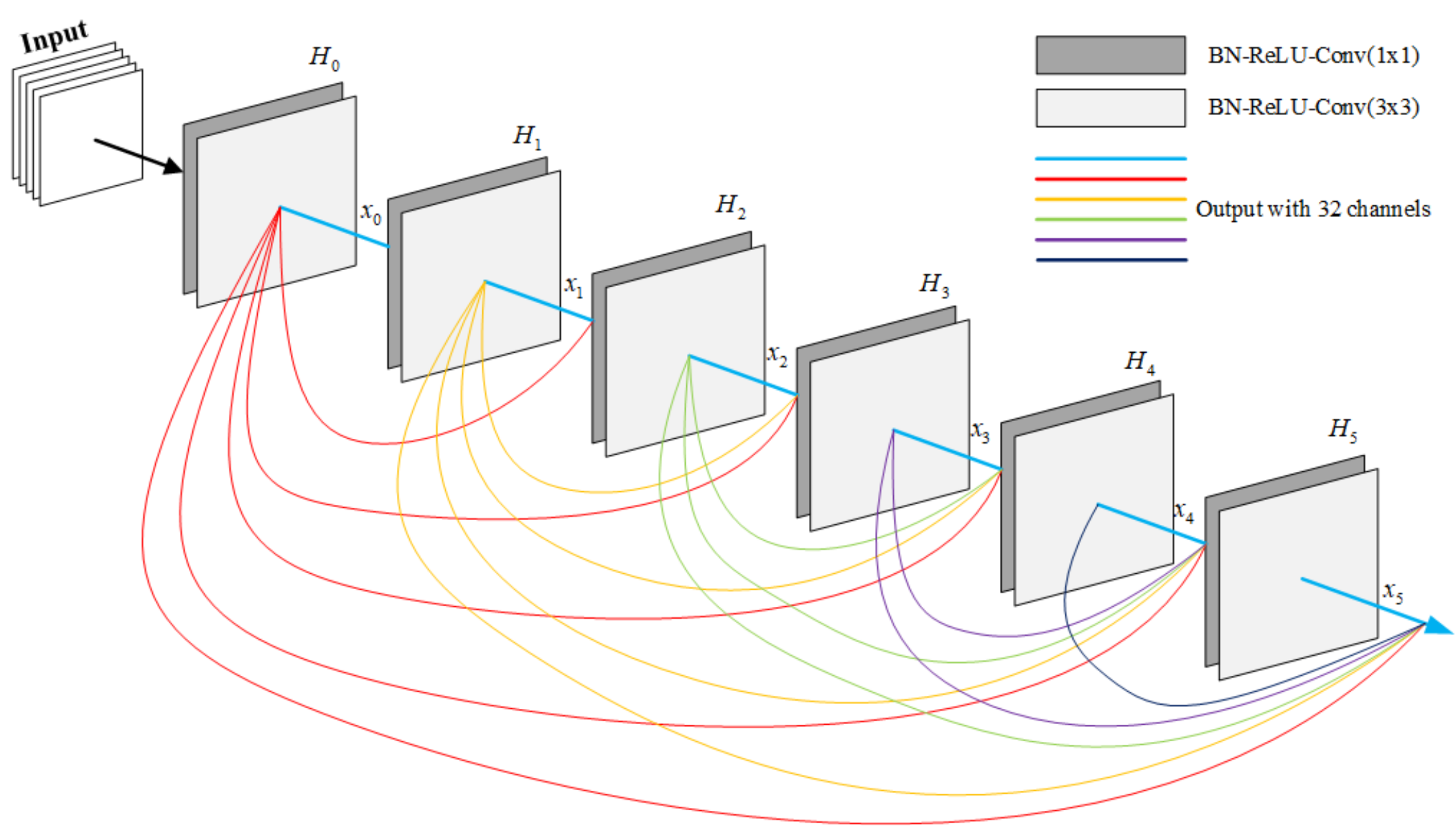}
			  \label{fig_block_structure:block}

	}
	
	\subfigure[Schematic structure of the network in our method.]{

			\centering
			\includegraphics[width=6.0in]{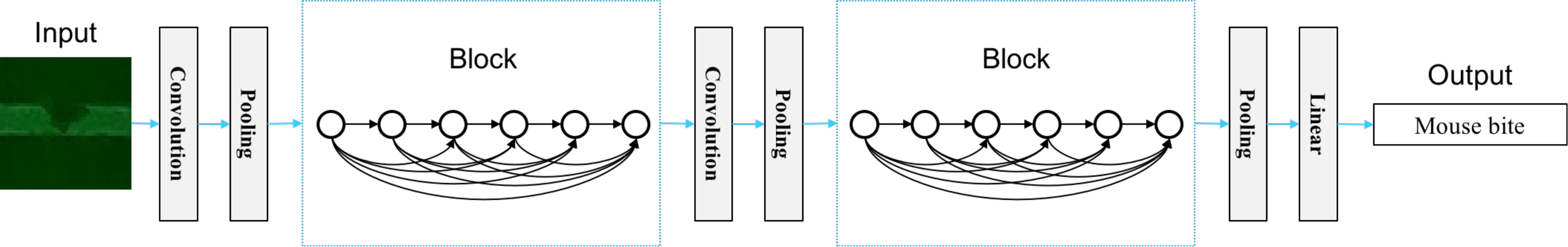}
			  \label{fig_block_structure:structure}

	}
	\caption{Basic Block and schematic network structure in our paper, (a) draws the 6-layer block and (b) illustrates the process on how output generates from an input image.}
	\label{fig_block_structure}
\end{figure*}

The network mainly consists of two basic blocks, as is illustrated in Figure \ref{fig_block_structure:block}. Each block has 6 convolutional layers, in which every layer takes all outputs of previous layers as input. Hence, the output of $l^{th}$ layer that has $l$ inputs (including outputs from previous block) can be defined as: 
\begin{equation}
\displaystyle
x_{l}=H_{l}([x_{0}, x_{1},...,x_{l-1}])
\end{equation}
where $x_{l}$ is the output of $l^{th}$ layer, $H_{l}(\cdot)$ denotes a compose of functions in $l^{th}$ layer including Batch Normalization (BN) \cite{Ioffe2015Batch}, Rectified Linear Units (ReLU) \cite{article_DSRNN}, Pooling \cite{article_GLADR}, and Convolution (Conv). In our experiment, each $H_{l}$ contains 2 convolutions of size $1 \times 1$ and siez $3 \times 3$ with stride 1 and padding 1 respectively, and there are BN and ReLU function before every convolution. The structure of a layer can be simply summarized as BN-ReLU-Conv ($1 \times 1$)-BN-ReLU-Conv ($3 \times 3$). Each $H_{l}$ is set to produce fixed 32 feature-maps, which will result in $l^{th}$ layer having $k_{0}+32\times(l-1)$ input feature-maps, here $k_{0}$ is the number of channels in the input layer. This $1 \times 1$ convolution can be introduced as bottleneck layer \cite{Szegedy2016Rethinking} to change number of input feature-maps, our method let each $1 \times 1$ convolution produce $ 4\times 32 $ feature-maps in block. More precisely, before sending into a layer, the feature-maps from previous layers are concatenated instead of combining them, so the 6-layer block will have 21 connections at last.  

In addition, before entering the first basic block, the input image will pass through a convolution of size $7 \times 7$ with stride 2 and padding 3, followed by BN, ReLU and Maxpooling function of size $3 \times 3$ with stride 2 and padding 1. Then the output will be passed to the first block which is followed by a transition layer where the number and size of feature-maps will be halved for compacting the model. The structure of transition layer is like BN-ReLU-Conv ($1 \times 1$)-AvgPool ($2 \times 2$). After the second block, an adaptive AvgPool is used and then a linear layer is employed to produce $6 \times 1$ vector. Detailed architecture of the network is demonstrated in Figure \ref{fig_block_structure:structure} and parameter setting is listed in Table \ref{table_architecture}.          

\subsection{Training}
The training process is executed on a computer with Intel Xeon E5-2640 CPU, 128GB RAM, and a NVIDIA GTX 1080Ti GPU is used during training. Stochastic gradient descent (SGD) with momentum 0.9 is used to update parameters. The initial learning rate is set to 0.01 and decay 0.1 every 7 epochs. We train the model using batch size 8 for 50 epochs, the whole training procedure takes about 25 minutes. L2 penalty is $ 1\times10^{-5} $ in the experiment to prevent over-fitting.      

\begin{table}
	\renewcommand{\arraystretch}{1.3}
	\caption{Detailed parameter setting of the network in our experiment.}
	\label{table_architecture}
	\centering
	\begin{tabular}{c|c|c}
		\hline
		Layers               & Output Size & Operation                               \\
		\hline
		Convolution          & $32\times32$   & $7\times7$ Conv, stride 2                  \\
		\hline
		Pooling              & $16\times16$   & $3\times3$ Max Pool, stride 2              \\
		\hline
  		Block1               & $16 \times 16$   & $\begin{bmatrix}
			$1$\times$1$ \quad Conv \\ 
			$3$\times$3$ \quad Conv 
		\end{bmatrix}$ $\times6$     \\
		\hline
		\multirow{2}*{Transition Layer}       & $16 \times 16$ & $1\times1$ Conv                            \\
		\cline{2-3}
		~                    & $8 \times 8$   & $2\times2$ Avg Pool                        \\
		\hline
		Block2               & $8\times8$         & $\begin{bmatrix}
			$1$\times$1$ \quad Conv  \\
			$3$\times$3$ \quad Conv 
		\end{bmatrix}$ $\times6$  \\
		\hline
		\multirow{2}*{Classification Layer} & $1 \times 1$   & $7 \times 7$ Adaptive Avg Pool             \\
		\cline{2-3}
		~                    & ~           & 6D fully-connected, softmax             \\
		\hline
	\end{tabular}
\end{table}   


\subsection{Results}
The goal of PCB defect inspection is defects detection and classification, while also minimizing the time expenditure of the method. Then, the performances of defect detection and defect classification are discussed. The metrics of defect detection is error rate $P_d$ that defined as follows: 
\begin{equation}
\displaystyle
P_d=\frac{\mid d-a\mid}{a}\times100\%
\end{equation}
where $d$ is the number of detected defect areas, $a$ is the actual number of defect areas. And the metrics of defect classification are the classification precision rate ($P_{c}$) of each type of defect and the average precision rate ($AP_{c}$). $P_{c}$ is defined in the following equation:
\begin{equation}
\displaystyle
P_c=\frac{c}{a}\times100\%
\end{equation}
in which $c$ is the correctly predicted number of a defect type, and $a$ is the actual number of defects of this type. And the average precision rate ($AP_{c}$) is defined as:
{\begin{equation}
\displaystyle
 AP_c=\frac{1}{N}\sum_{i=1}^NP_c^i
\end{equation}
where $P_c^i$ is the precision rate of $i^{th}$ defect, $N$ denotes the number of types of defects, which is 6 in this paper.

\subsubsection{Defects inspection}
To verify the effectiveness of our reference comparison based method, we implemented the preprocessing and detection algorithm on our dataset. The statistics of the result is listed in Table \ref{table_defects_det}. We can see that only a mouse bite and open circuit defect are needlessly detected， the former is a wrong detection (ie, false detection of non-defect area) and the latter is an overlapped one (ie, a defect produces two similar overlapped results).

\begin{table*}
	\centering
	\begin{minipage}{1.0\textwidth}

		\renewcommand{\arraystretch}{1.3}
		\caption{Defects detection results, first row lists number of defects provided by dataset, second row lists number of defects got by our method, and third row are the error rate P\_{d}s.}
		\label{table_defects_det}
		\centering
			\begin{tabular}{lllllll}
						\hline
						~               & Missing hole & Mouse bite    & Open circuit        & Short & Spur & Spurious copper \\
						\hline
						Actual number   & 497          & 492           & 482                 & 491   & 488  & 503             \\
						\hline
						Detected number & 497          & 493(+1 error) & 483(+1 over lapped) & 491   & 488  & 503             \\
						\hline
						Error rate (P\_d)		& 0\%          & 0.2\% & 0.2\% & 0\%    & 0\%   & 0\%              \\
						\hline
					\end{tabular}

	\end{minipage}
	\hspace{4in}
	\begin{minipage}{1.0\textwidth}
	\renewcommand{\arraystretch}{1.3}
	\caption{Defects classification results, the first row represents the $P_c$s and $AP_c$ obtained in the test data produced in Section \uppercase\expandafter{\romannumeral4} (A), and the second row represents the $P_c$s and $AP_c$ obtained from all samples in Section \uppercase\expandafter{\romannumeral3}.}
	\label{table_classification}
	\centering
	\begin{tabular}{llllllll}
					\hline
					~            & Missing hole & Mouse bite & Open circuit & Short   & Spur    & Spurious copper & Average (AP\_c) \\
					\hline
					Test data (P\_c) & 98.96\%        & 97.94\%     & 97.74\%      & 99.48\% & 93.65\% & 98.52\%         & 97.74\% \\
					\hline
					All samples (P\_c) & 100\%        & 99.6\%     & 99.18\%      & 99.39\% & 99.39\% & 98.80\%         & 99.40\% \\
					\hline
				\end{tabular}
	\end{minipage}
\end{table*}

\subsubsection{Defects classification}
We test our classification model on the test data produced in section \uppercase\expandafter{\romannumeral4} (A) by bounding box and all defect samples produced in section \uppercase\expandafter{\romannumeral3} by our reference comparison based method. It should be noted that before classifying the defects, we remove the repeatedly and incorrectly detected samples in the defect detection results to avoid the impact on the classification procedure. The result showed in Table \ref{table_classification} indicates that our method acquire superior performances on both groups, with average precision of 97.74\% and 99.40\%, respectively. The reason for this case is that original defect image obtained by the reference comparison method is smaller than the image cropped by the bounding box given by the manual annotation. Resulting in the defect body accounts for a larger proportion when the image is resized to fixed resolution, which is more beneficial for classification.


\subsubsection{Time consumption}
Taking the detection efficiency into account, we recorded the time required to spend in each step of inspecting a PCB, as described in Table \ref{table_time_consumption}. It takes a total of 0.9899 second to execute the entire process on a computer with Intel Core i7-7700 CPU @ 3.60GHz, 8GB RAM. In these steps, registration accounts for the most of total time because searching feature points and calculating descriptors are all time consuming tasks.

\begin{table}[!t]
	\renewcommand{\arraystretch}{1.3}
	\caption{Steps of filtering and mathematical morphology processing in our experiment.}
	\label{table_step_fil_mm}
	\centering
	\begin{tabular}{l|l|l}
		\hline
		Step & Operation        & Kernel(shape)           \\
		\hline
		1    & median filtering & 5x5     \\
		\hline
		2    & closing          & 15x15(rectangle) \\
		\hline
		3    & opening          & 3x3(rectangle)   \\
		\hline
		4    & median filtering & 5x5      \\
		\hline
		5    & closing          & 29x29(ellipse)   \\
		\hline
		6    & opening          & 3x3(rectangle)   \\
		\hline
		7    & opening          & 1x1(rectangle)   \\
		\hline
	\end{tabular}
\end{table}

\begin{table}[!htbp]
	\renewcommand{\arraystretch}{1.3}
	\caption{Time consumption of each step, registration takes up the most time, followed by localization and binaryzation.}
	\label{table_time_consumption}
	\centering
	\begin{tabular}{c|c}
		\hline
		Procedure      & Time(s) \\
		\hline
		Registration   & 0.6219  \\
		\hline
		Binaryzation   & 0.1650  \\
		\hline
		Localization   & 0.1808  \\
		\hline
		Classification & 0.0212  \\
		\hline
		Total & 0.9889  \\
		\hline
	\end{tabular}
\end{table}   

\section{Conclusion}
In this paper, in consideration of lack of public shared PCB dataset, we produce and publicize a synthesized PCB dataset that has 1386 images with 6 types of common defects, including missing hole, mouse bite, open circuit, short, spur, spurious copper. Half of the images are for the situation where a test PCB is placed correctly, while other half is set for simulating the situation when the test board is randomly orientated in the workbench. Bounding box of every defect are provided in our dataset so that the location of each defect can be affirmed, besides, the existing of bounding box makes it possible for the images to be utilized as labeled data in object detection tasks. The transformation information is also provided to facilitate other researchers to study registration problems.     

Based on reference comparison method, we introduce an end-to-end convolutional neural network model to classify the defects, which reaches impressive performance on our dataset. In order to learn more effectively, we do not choose simply stacking convolutional layers, instead, we use dense shortcuts inspired from Densenet to achieve high accuracy with relatively few layers. 

Future work may focus on continuously increasing the size of the dataset, improving the robustness of the algorithm, reducing the time consumed of the entire detection process while achieve higher efficiency, what's more , designing effective non-reference compaison method to avoid using template.


%

%
%
%
%
\section*{Acknowledgment}

The authors would like to thank...

\ifCLASSOPTIONcaptionsoff
  \newpage
\fi



\bibliographystyle{plain}
\bibliography{paper}
\end{document}